# Impact of Ground Truth Annotation Quality on Performance of Semantic Image Segmentation of Traffic Conditions

Vlad Taran, Yuri Gordienko, Alexandr Rokovyi, Oleg Alienin, Sergii Stirenko

National Technical University of Ukraine "Igor Sikorsky Kyiv Polytechnic Institute", Kyiv, Ukraine
`vladtkv@gmail.com`

**Abstract.** Preparation of high-quality datasets for the urban scene understanding is a labor-intensive task, especially, for datasets designed for the autonomous driving applications. The application of the coarse ground truth (GT) annotations of these datasets without detriment to the accuracy of semantic image segmentation (by the mean intersection over union — mIoU) could simplify and speedup the dataset preparation and model fine tuning before its practical application. Here the results of the comparative analysis for semantic segmentation accuracy obtained by PSPNet deep learning architecture are presented for fine and coarse annotated images from Cityscapes dataset. Two scenarios were investigated: scenario 1 — the fine GT images for training and prediction, and scenario 2 — the fine GT images for training and the coarse GT images for prediction. The obtained results demonstrated that for the most important classes the mean accuracy values of semantic image segmentation for coarse GT annotations are higher than for the fine GT ones, and the standard deviation values are vice versa. It means that for some applications some unimportant classes can be excluded and the model can be tuned further for some classes and specific regions on the coarse GT dataset without loss of the accuracy even. Moreover, this opens the perspectives to use deep neural networks for the preparation of such coarse GT datasets.

## 1      Introduction

The current tendency in semantic image segmentation of traffic road conditions is making high quality images labeling to produce fine ground truth (GT) annotations for training and testing deep learning networks [1-3]. That stage of labeling is especially difficult, as it can take several hours to fine annotate single image. Considering the fact, that autonomous driving system can be used in different cities or regions, the model tuning is necessary to achieve good segmentation accuracy. Despite the availability of the proprietary (like Daimler AG [4], Tesla [5], etc.) and open-source datasets of traffic road conditions (like Cityscapes [6], KITTI [7], CamVid [8-9], DUS [10], etc.), the lack of sample datasets for specific application regions hardens the model tuning stage. In the context of self-driving cars, for the some components like automatic braking and anti-collision systems, it may be not necessary to annotate all possible object classes for semantic segmentation tasks. It could be enough to select main classes such as a road, a car, a person/pedestrian, traffic lights, and traffic signs. It could also not necessary to make high quality annotations of the traffic condition while preparing training or testing datasets. It is assumed that coarse objects shapes may be suitable for the common object identification and localization on the road. Such a coarse annotation could speedup a datasets preparation for the model adaptation to the different use cases and model fine tuning before its practical application. But the problem is that the impact of decreasing ground truth annotation quality on performance of semantic image segmentation of traffic conditions was not investigated yet, especially for different classes and in the presence of the specific application regions, for example, for cities and countries with various architecture and urban life styles.

The main aim of this paper is to investigate the accuracy of semantic image segmentation by classes and the impact of the GT annotation quality on the performance of semantic image segmentation in the context of urban scene and traffic conditions understanding. The section *2.Background and Related Work* gives the brief outline of the state of the art in urban scene datasets, networks used, and specific problems of dataset annotation. The section *3.Experimental and Computational Details* contains the description of the experimental part related with the selected dataset, network, scenario, and metrics used. The section *4.Results* reports about the experimental results obtained, the section *5.Discussion* is dedicated to the discussion of these results, and section *6.Conclusions* summarizes the lessons learned.

## 2     Background and Related Work

Currently, the street scenes are typically monitored by multiple input modalities and the obtained results are represented in the various relevant datasets:
- a stereo camera, traffic light camera, localization camera — Daimler AG - Research & Development proprietary solution and dataset [4];
- 8 surround cameras (in addition to 12 ultrasonic sensors and forward-facing radar; the radar can see vehicles through heavy rain, fog or dust) — Tesla Autopilot proprietary solution and dataset [5].
- a stereo camera rig that gives a diverse set of stereo video sequences recorded in street scenes from 50 cities, with high quality pixel-level annotations of 5 000 frames in addition to a larger set of 20 000 weakly annotated frames, — Cityscapes public dataset (https://www.cityscapes-dataset.com/) [6];
- two stereo camera rigs (one for grayscale and one for color), 3D laser scanner, GPS measurements and IMU accelerations from a combined GPS/IMU system, timestamps — KITTI public dataset [7];
- traffic cam videos — the open access image database Cambridge-driving Labeled Video Database (CamVid) (ftp://svr-ftp.eng.cam.ac.uk/pub/eccv/) [8-9];
- stereo image pairs — Daimler Urban Segmentation (DUS) public dataset [10]

Recently, several new networks appeared, like PSPNet[11], ICNet[12], DeepLab [13] and many others, and demonstrated the high performance with regard to the accuracy of semantic image segmentation (by the mean intersection over union — mIoU) and speed of prediction (by the inference time) of various object classes, for example, for some of these datasets.

In addition to this, there is the comparative analysis of ICNet and PSPNet performance with regard to several subsets of Cityscapes dataset including stereo-pair images taken by left and right cameras for different cities [14]. It was found that the distributions of the mIoU values for each city and channel are asymmetric, long-tailed, and have many extreme outliers, especially for PSPNet network in comparison to ICNet network. The results obtained demonstrated the different sensitivity of these networks to: (1) the local street view peculiarities (among different cities), (2) the change of viewing angle on the same street view image (right and left data channels). The differences with regard to the local street view peculiarities should be taken into account during the targeted fine tuning the models before their practical applications. For both networks, the information from the additional right data channel is radically different from the left channel, because it is out of the limits of statistical error in relation to the mIoU values. It means that the traffic stereo pairs can be effectively used not only for depth calculations (as it is usually used), but also as an additional data channel that can provide much more information about scene objects than a simple duplication of the same street view images.

In the view of the aforementioned background and the different quality of annotated images in various datasets, the aim of this work stated in section 1.Introduction was formulated as impact study of ground truth annotation quality on performance of semantic image segmentation of traffic conditions. That is why we have tried to estimate the difference between accuracy values while applying fine and coarse annotated ground truth images.

## 3     Experimental and Computational Details

Here PSPNet network was used as semantic image segmentation model. It took the first place in ImageNet [15] scene parsing challenge in 2016, PASCAL VOC [16] and Cityscapes semantic segmentation benchmarks. The goal of this network was to extract complex scene context features. While solving the several common issues such as mismatched relationship, confusion categories and inconspicuous classes, it introduce the state-of-art pyramid pooling module which extend the pixel-level features to make the final prediction more reliable. This module separates the feature map into four different pyramid scales and form the pooled representation for different locations. It allows collecting levels of information, more representative than the global pooling.

For this investigation images from Cityscapes dataset were used that correspond to several German cities like Frankfurt (267 images), Munster (174 images), Lindau (59 images) with fine GT (Fig. 1), and Erlangen (265), Konigswinter (118), Troisdorf (138) with coarse GT (Fig. 2).

Because of the large amount of the recorded data, Cityscapes dataset provide fine and coarse annotated images. The fine image annotation means pixel-precision annotation for 30 classes for 3 cities, containing 5 000 images in total. The coarse image annotation means line-precision annotation for the remaining 23 cities, containing 20 000 images in total, where a single image was selected every 20 s or 20 m driving distance and spending less than 7 min of the annotation time per image [17].

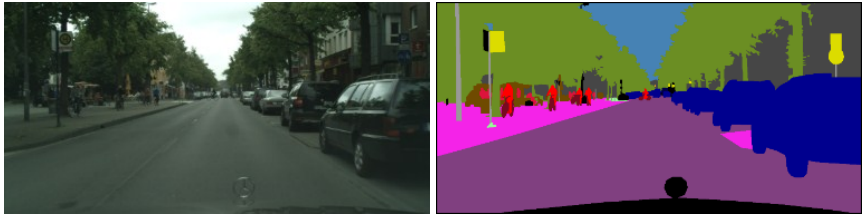

**Fig. 1.** Example of the original image from Cityscapes dataset (left) and its fine ground truth with pixel-wise precision (right) for Munster.

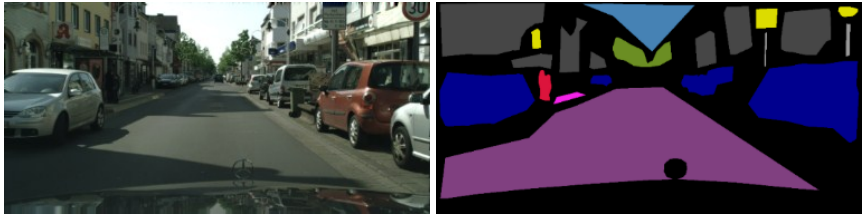

**Fig. 2.** Example of the original image from Cityscapes dataset (left) and its coarse ground truth with line-bounded precision (right) for Troisdorf.

PSPNet network, pre-trained on Cityscapes dataset, was applied to these data subsets to measure the image prediction accuracy (mIoU) and compare them for fine and coarse ground truth annotations for the corresponding cities. The whole workflow was carried out according to the scenarios 1 and 2 described in the Table 1, namely, for scenario 1 the fine GT images were used for training and prediction, and for scenario 2 the fine GT images were used for training and the coarse GT images were used for prediction. It should be noted that scenario 1 was already implemented in our previous work [14] and scenarios 3 4 are under work right now and will be published elsewhere [18]. In this, work the following main classes, which could be used in the autonomous driving tasks, were selected: a road, a car, a person/pedestrian, traffic lights and signs classes.

**Table 1.** Scenarios for comparison of impact of ground truth annotation quality.

| Scenario | Training | Prediction | Implementation |
|---|---|---|---|
| 1 | Fine | Fine | [14], this work, [18] |
| 2 | Fine | Coarse | this work, [18] |
| 3 | Coarse | Fine | [18] |
| 4 | Coarse | Coarse | [18] |

All experiments were conducted on the basis of TensorFlow framework [19] on the workstation with the single NVIDIA GTX 1080 Ti GPU card with CUDA 8.0 and CUDNN 7.

## 4    Results

The accuracy of semantic image segmentation was measured by the mean intersection over union (mIoU) parameter. It was used to compare the per-class prediction performance for images with the various segmentation quality (fine and coarse GT annotations) for the following classes: a road, a car, a person/pedestrian, traffic lights and signs. The results of accuracy measurements obtained by PSPNet network for different classes and quality of semantic image segmentation are shown for the corresponding cities in Fig. 3.

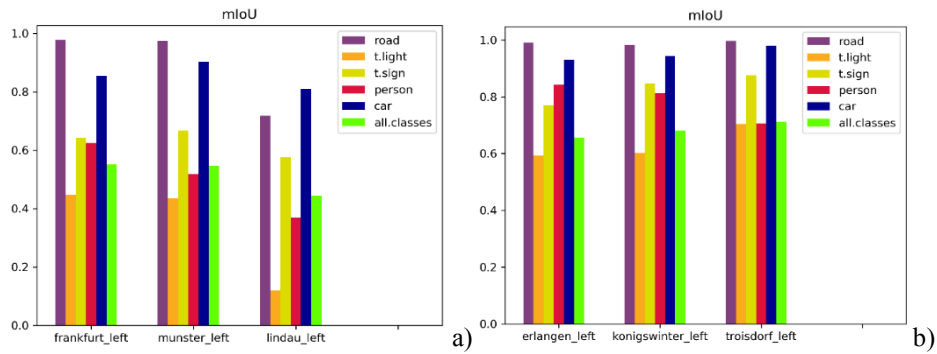

**Fig. 3.** Per-class accuracy (mIoU) values for each city with fine (a) and coarse (b) GT obtained by PSPNet network.

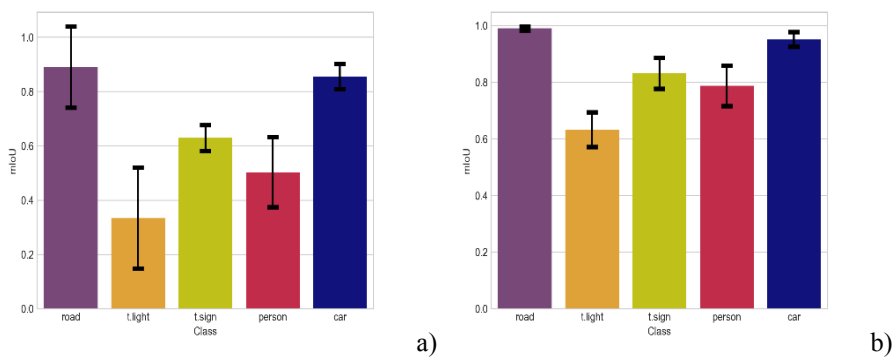

**Fig. 4.** Mean and standard deviations (shown by whiskers) of per-class accuracy (mIoU) values for all cities with fine (a) and coarse (b) GT.

The mean and standard deviations of per-class accuracy (mIoU) values for all cities are shown in Fig. 4 and listed in Table 2 for fine (Fig. 4a) and coarse (Fig. 4b) GT.

**Table 2.** The mean and standard deviations (shown by whiskers) of per-class accuracy (mIoU) values for all cities (Fig. 4).

| Object Class | Mean | | Standard deviation | |
|---|---|---|---|---|
| | Fine | Coarse | Fine | Coarse |
| **Road** | 0.89 | 0.99 | 0.14 | 0.007 |
| **Car** | 0.85 | 0.95 | 0.04 | 0.02 |
| **Traffic sign** | 0.62 | 0.83 | 0.05 | 0.05 |
| **Person** | 0.5 | 0.78 | 0.12 | 0.07 |
| **Traffic light** | 0.33 | 0.63 | 0.19 | 0.06 |

The common tendency is that the means of per-class accuracy (mIoU) for all cities with fine GT are lower than for the coarse GT, and it is vice versa for the standard deviations of mIoU.

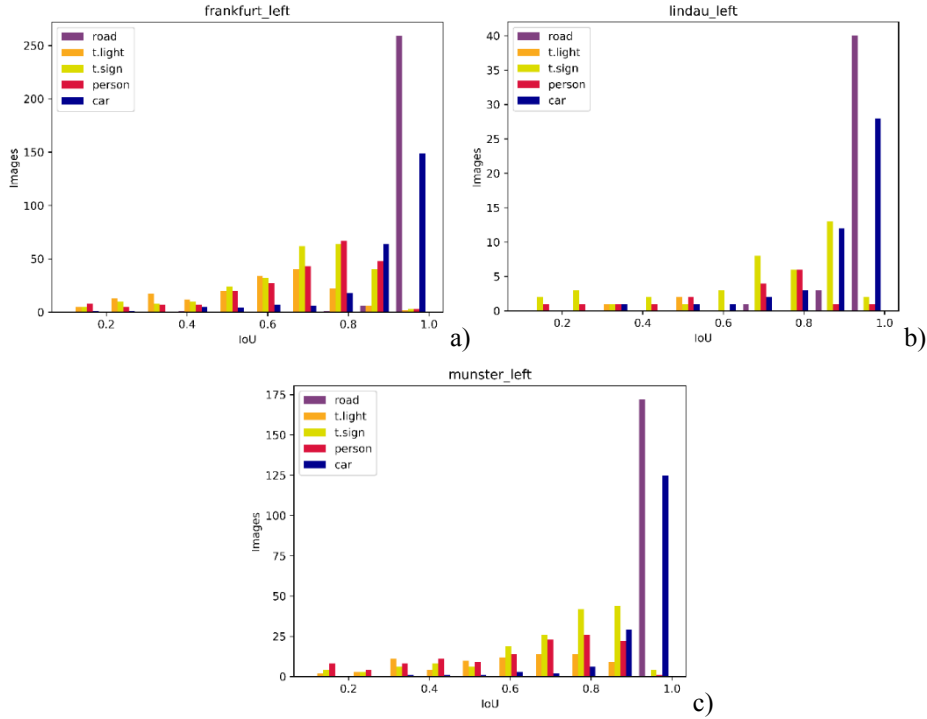

**Fig. 5.** Distributions of accuracy values (mIoU) for set of left images with fine ground truth from cities: (a) Frankfurt, (b) Lindau and (c) Munster.

The presented plots for fine and coarse GT allow us to make their comparison with regard to the accuracy (mIoU). Variations of mIoU (Fig. 3) among classes with fine and coarse GT can be very big, for example, for fine GT (Fig. 3a) the mIoU was equal to 0.5±0.07, and the worst segmented was the traffic light class with the mIoU equal to 0.33±0.19. The best segmented classes were the road and car ones, with the mIoU equal to 0.89±0.14 and 0.85±0.04, respectively (Fig. 4a, Table 2). The mIoU for the traffic sign class was equal to 0.62±0.05, and for the person class it was equal to 0.5±0.12, except for Lindau city, where it was equal to 0.38.

As to the coarse GT, the mIoU was greater than 0.6 for all classes (Fig. 3b). The road and car classes were the most accurately segmented ones with the mIoU equal to 0.99±0.007 and 0.95±0.02, respectively (Fig. 4b, Table 2). The person class accuracy was equal to 0.78±0.07, except for Troisdorf city, where it was equal to 0.71 (Fig. 3b). The traffic light and sign classes have accuracy equal to 0.63±0.06 and 0.83±0.05 respectively.

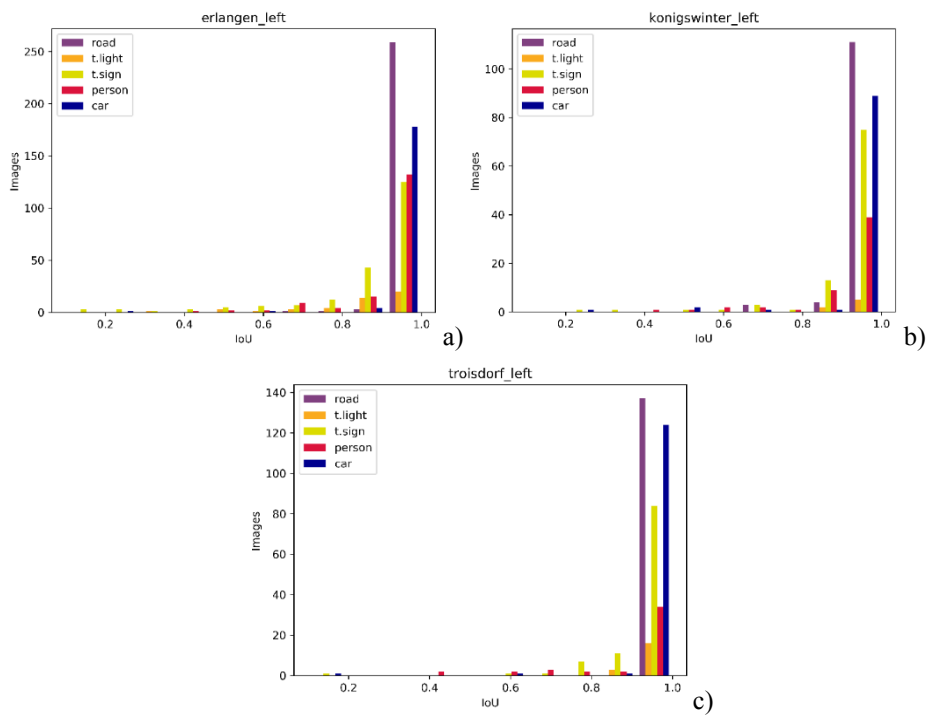

**Fig. 6.** Distributions of accuracy values (mIoU) for set of left images with coarse ground truth from cities: (a) Erlangen, (b) Konigswinter and (c) Troisdorf.

The distributions of the IoU values were determined for each type of GT annotation: fine — (Frankfurt, Munster, Lindau); coarse – (Erlangen, Konigswinter, Troisdorf), and presented in Fig. 5 and Fig. 6, respectively.

The qualitative analysis of distributions for the fine GT allows us to make the assumptions that they are asymmetric, and long-tailed (Fig. 5). In contrast, the distributions for coarse ground truth are short-tailed and have fewer outliers (Fig. 6).

## 5   Discussion

The results obtained allow us to compare the accuracy values of semantic image segmentation for images with different GT annotations: fine and coarse. While coarse annotated images are less detailed compared with fine annotated images, they still allowed us to identify, localize, and segment rough shapes of the objects from the preselected classes in the context of urban scenes. It is assumed that it can be suitable for autonomous car task where the object localization may be much more important than getting the exact shape of the object.

In general, the IoU values for coarse GT annotated images were greater than for the fine GT annotated ones. It can be explained by the assumption that objects in coarse GT images have more rough shapes (than in the fine GT images) that are more close to the correspondent pre-trained primitives inside the model itself and the accuracy is measured regarding their overlapping. In reverse, the objects in fine GT images have more detailed shapes (than in coarse GT images) that are far from to the correspondent pre-trained primitives inside the model itself.

But it should be realized that the presented results are based on the coarse segmentation analysis for only three cities from Cityscapes dataset. Therefore, the reliability of these results for other cities and countries should be verified by the wider range of images on other cities and countries that is under work right now. In addition, the current model was trained on Cityscapes dataset with fine GT annotations, namely for scenarios 1 and 2 (Table 1). And it is worth to estimate the segmentation accuracy for scenario 3 (Table 1), i.e. for the cases of training on coarse GT annotated images and prediction on fine GT images, and scenario 4 (Table 1), i.e. for the cases of training on coarse GT annotated images and prediction on coarse GT images. But these results are under work yet and will be reported in details elsewhere [18].

Considering the fact that creating quality datasets for urban scene understanding is a labor-intensive task, especially, for datasets designed for autonomous driving applications and model fine tuning before their practical applications, we proposed to limit object classes to only necessary ones. The further question can be raised about excluding some irrelevant classes (like a sky, a building, a vegetation and sidewalk) in some applications like emergency braking and anti-collision systems [20-22] and other applications [23-26].

The accuracy comparison for fine and coarse GT annotated images allow us to raise the question that deep neural networks may be used for creation of coarse GT annotated datasets which can be edited and used for fine tuning the pre-trained models for the specific application regions.

The additional promising development of this work could be related with the more rational segmentation where an object shape could be segmented not in the fine pixelwise way, but by the means of only specific coarse features using lines, curves, arcs, etc. For example, such rational per-class segmentation for the road could be implemented by complex lines or arcs, for cars — by polygons, for traffic signs — by geometric primitives like ellipses and polygons, etc. But for the complex objects like pedestrians, some optimal methods should be investigated to distinguish moving people, as potentially more dangerous, and standing ones.

## 6     Conclusions

The comparative analysis of semantic segmentation accuracy was performed in this work. The results were obtained by PSPNet deep learning architecture for fine and coarse annotated images from Cityscapes dataset. The four possible scenarios of training/testing on fine/coarse GT images were proposed and two of them were considered: scenario 1 — the fine GT images for training and prediction, and scenario 2 — the fine GT images for training and the coarse GT images for prediction.

The results shown that for the most important classes (road, car, person/pedestrian, traffic light, traffic sign) mean accuracy values of semantic image segmentation for coarse GT annotations are higher than for fine GT ones, and the standard deviations are vice versa. Despite the coarse annotated images are less detailed compared with fine annotated ones, they still allowed us to identify, localize, and segment the rough shapes of the objects from the preselected classes in the context of urban scenes. Moreover, it was possible without detriment to the accuracy of semantic image segmentation (mIoU). It means that for some applications (like emergency braking and anti-collision systems) some unimportant classes (like sky, building, vegetation, sidewalk, etc.) can be excluded and the model can be tuned further for some classes and specific regions on the coarse GT dataset without loss of the accuracy even. Moreover, this opens the perspectives to use deep neural networks for the preparation of such coarse GT datasets, and the usage of the more rational segmentation where an object shape could be segmented not in a fine pixel-wise way, but by the means of only specific coarse features using lines, curves, arcs, etc.